# MRI-Based Brain Tumor Detection through an Explainable EfficientNetV2 and MLP-Mixer-Attention Architecture

*Mustafa YURDAKUL[1*]*  *, Şakir TAŞDEMİR[2]*

[1*]Kırıkkale University, Computer Engineering Dept, Kırıkkale, Türkiye, mustafayurdakul@kku.edu.tr
[2]Selçuk University, Computer Engineering Dept, Konya, Türkiye, stasdemir@selcuk.edu.tr

**Abstract:**

**Background and objective**

Brain tumors are serious health problems that require early diagnosis due to their high mortality rates. Diagnosing tumors by examining Magnetic Resonance Imaging (MRI) images is a process that requires expertise and is prone to error. Therefore, the need for automated diagnosis systems is increasing day by day. In this context, a robust and explainable Deep Learning (DL) model for the classification of brain tumors is proposed.

**Methods**

In this study, a publicly available Figshare dataset containing 3,064 T1-weighted contrast-enhanced brain MRI images of three tumor types was used. First, the classification performance of nine well-known CNN architectures was evaluated to determine the most effective backbone. Among these, EfficientNetV2 demonstrated the best performance and was selected as the backbone for further development. Subsequently, an attention-based MLP-Mixer architecture was integrated into EfficientNetV2 to enhance its classification capability. The performance of the final model was comprehensively compared with basic CNNs and the methods in the literature. Additionally, Grad-CAM visualization was used to interpret and validate the decision-making process of the proposed model.

**Results**

The proposed model's performance was evaluated using the five-fold cross-validation method. The proposed model demonstrated superior performance with 99.50% accuracy, 99.47% precision, 99.52% recall and 99.49% F1 score. The results obtained show that the model outperforms the studies in the literature. Moreover, Grad-CAM visualizations demonstrate that the model effectively focuses on relevant regions of MRI images, thus improving interpretability and clinical reliability.

**Conclusion**

A robust deep learning model for clinical decision support systems has been obtained by combining EfficientNetV2 and attention-based MLP-Mixer, providing high accuracy and interpretability in brain tumor classification.

**Keywords;** Attention, Brain Tumor Classification, Magnetic Resonance Imaging, Explainable Artificial Intelligence, MLP-Mixer, EfficientNetV2

# 1. Introduction

Brain tumor is a benign or malignant mass formed by the uncontrolled and abnormal growth of cells in the brain tissue[1, 2]. Today, brain tumors affect hundreds of thousands of people worldwide every year, have high mortality rates, especially with malignant types, and are among the most deadly diseases among central nervous system tumors. In 2024, approximately 25,000 new cases of brain tumors were diagnosed in the USA and a significant proportion of these cases resulted in death[3].

The most common types of brain tumors encountered clinically are glioma, meningioma and pituitary tumors[4].

Glioma originates from glial cells, the support cells of the brain. Structurally heterogeneous, a glioma is usually a rapidly progressing and aggressive tumor[5, 6]. Meningioma is a tumor that develops from the meninges, the membrane tissue surrounding the brain and spinal cord[7]. It is usually benign and tends to grow slowly[8]. However, it is possible for a meningioma to cause various neurological problems by pressing on the brain tissue as it grows. Pituitary Tumor occurs in the pituitary gland, which is one of the important endocrine structures of the brain. It is classified as functional and non-functional according to its hormone secretion status[9]. While functional pituitary tumors affect the hormone levels in the body and cause serious clinical symptoms[10], non-functional pituitary tumors usually cause symptoms that occur as a result of pressure on surrounding tissues[11]. MRI is one of the most common imaging modalities for detecting brain tumors[12, 13]. MRI provides soft tissue contrast with high resolution, allowing detailed examination of the location and size of tumors. However, the analysis of MRI images is a time-consuming and complex process that can vary depending on the interpreter[14]. In particular, the morphological diversity of tumors, difficulties in distinguishing similar tissue densities, and the possibility of missing small lesions are limitations of manual evaluation. Therefore, the need for computer-aided diagnosis (CAD) systems is becoming more and more important. There are various approaches to detect brain tumors in the literature.

Swati et al.[15] proposed a convolutional neural network (CNN) model trained from scratch for the classification of brain tumors. The Figshare dataset of T1-weighted contrast-enhanced MRI images was used in the study, and a total of 3064 images were classified into three tumor classes (glioma, meningioma, pituitary). The model architecture consists of five convolution-pool blocks. The classification performance achieved an accuracy of 96.49%, 92.79% precision, 96.33% recall and 94.47% f1 score on the test set.

Aamir et al. [16] proposed a Deep Learning(DL) based approach to classify brain tumors using MRI images. In order to improve image quality, illumination boost and non-linear stretching were applied. Then, features extracted with EfficientNet-B0 and ResNet50 models are combined with the Partial Least Squares (PLS) method. Tumor regions were determined with agglomerative clustering and classification was performed with ROI pooling. The study was performed on the Figshare dataset and 98.95% accuracy rate was achieved.

Ismail et al. [17] used the ResNet50 deep learning model in their study conducted on the same dataset. They utilized various data preprocessing steps, data augmentation techniques, and transfer learning approaches to improve the model's performance. According to the results reported in the study, 97% accuracy, 98% precision, 97% recall, and an F1-score of 97% were achieved.

Hashemzehi et al. [18] proposed a hybrid model combining NADE and CNN methods. In this method, convolutional layers were used instead of fully connected layers. Their architecture consists of two separate networks, and the final layers of these networks are connected to the output layer. The authors achieved recall, precision, and F1 score values of 94.64%, 94.49%, and 94.56%, respectively, with their proposed approach on the same dataset.

Sharma et al. [19] proposed a transfer learning based deep learning method for brain tumor classification. In the study, the ResNet50 model, customized by removing the last layer, was used on an other MRI image dataset containing glioma, meningioma, pituitary and normal classes. The performance of the model achieved 92% accuracy, 89% recall and 93% specificity in tumor type classification.

Khaliki et al. [20] evaluated different CNN architectures (Inception-V3, EfficientNetB4 and VGG19) on a total of 2870 brain MRI images including four classes: glioma, meningioma, tumor-free and pituitary. In the experiments, the highest performance was achieved by VGG16 with 98% accuracy, 97% F-score, 98% recall and 98% precision.

Khan et al. [21] proposed a novel lightweight CNN model called ShallowMRI. The model utilizes an efficient erosion/dilation-based contour extraction and Local Binary Pattern (LBP) features, as well as an novel attention mechanism. Experiments on the Kaggle MRI brain tumor dataset yielded 98.24% accuracy, 98.17% precision, 98.10% recall and 98.13% F1 score.

Shaikh et al. [22] used stacking ensemble learning(SEL) model for tumor detection and segmentation, combining the outputs of different CNN-based models (3D-CNN, MobileNet-v3, VGG-16, VGG-19, ResNet50, AlexNet) with DenseNet201 designed as a meta-model. Experiments on the BraTS dataset show that the SEL model achieved very high performance with 99.65% accuracy, 99.76 % precision, 98.97% recall and 98.99% F1 Score.

Davar et al. [23] proposed a two-stage method for brain tumor segmentation on the Figshare dataset, incorporating YOLOv3 and U-Net architectures enhanced with spatial-channel attention modules. In the first stage, regions of interest (ROIs) of the tumor were detected using YOLOv3, and in the second stage, tumors in these regions were accurately segmented using U-Net. The study achieved 91.27% precision, 89.74% recall, 92.73% Jaccard index, and 89.15% Dice coefficient.

**Table 1.** Summary of recent literature studies on brain tumor classification

| Study | Method | Results | Limitations |
|---|---|---|---|
| Swati et al. (2019) | Custom CNN model | Accuracy: 96.49%, Precision: 92.79%, Recall: 96.33%, F1 Score: 94.47% | The simple CNN structure has limited learning of complex tumor patterns. |
| Ismael et al. (2020) | ResNet50 | Accuracy:97% Precision:98% Recall:97% F1-Score:97% | The simple CNN structure has limited learning of complex tumor patterns. |
| Hashemzehi et al. (2020) | NADE + CNN | Recall: 94.64% Precision: 94.49% F1 Score: 94.56% | Relatively low performance and complex structure |
| Aamir et al. (2022) | EfficientNet-B0 & ResNet50, feature fusion with PLS, agglomerative clustering | Accuracy: 98.95% | The multi-stage structure and high computational cost are limiting in practice |
| Sharma et al. (2023) | ResNet50 | Accuracy: 92%, Recall: 89%, Specificity: 93% | The explainability of the model is low; the decision process is not clinically interpretable. |
| Khan et al. (2024) | VGG16 | Accuracy: 98% F Score: 97% Recall: 98% Precision: 98% | The explainability of the model is low; the decision process is not clinically interpretable. |
| Khan et al. (2025) | ShallowMRI | Accuracy: 98.24%<br>Precision: 98.17%<br>Recall: 98.10%<br>F1 score: 98.13% | The success of the model is highly dependent on preprocessing steps, which may or may not be consistent across different data qualities or imaging protocols |
| Shaikh et al. (2025) | SEL model | Accuracy: 99.65%<br>Precision: 99.76 %<br>Recall: 98.97%<br>F1 Score: 98.99% | The SEL model has high computational complexity and limited explainability. |
| Davar et al.(2025) | YOLOv3 + U-Net | Precision: 91.27%,<br>Recall: 89.74%,<br>Jaccard: 92.73%,<br>Dice score: 89.15%. | Relatively low performance |

Table 1 summarizes the representative studies in the literature on brain tumor classification, highlighting their methods, performance results, and limitations. When the studies summarized in Table 1 are examined, it can be seen that different CNN-based deep learning approaches have been developed for brain tumor detection in recent years. Although the studies have generally achieved high accuracy rates, the high computational cost, complex structure, or limited clinical interpretability of most models stand out as major drawbacks. In light of the achievements and limitations reported in the literature, this study presents a novel hybrid approach that balances high performance with clinical interpretability.

The main contributions of this study can be summarized as follows:
- A comprehensive and comparative literature review was conducted on the detection of brain tumors using artificial intelligence approaches.
- The Figshare dataset, which is widely used in brain tumor classification studies and includes different tumor types, was used as the dataset for this research.
- Nine different CNN-based architectures with different feature extraction blocks were evaluated on the dataset.
- A novel and robust hybrid model combining EfficientNetV2 with an attention-based MLP-Mixer architecture was proposed. The performance of the proposed model was compared with both basic CNN models and recent literature studies.
- Gradient-weighted Class Activation Mapping(Grad-CAM) visualizations were applied to explain the model's decision-making process for clinical interpretability.

The remainder of this paper is structured as follows. Section 2 describes the dataset, the proposed methodology, and the attention-based MLP-Mixer architecture. Section 3 presents the experimental setup and evaluation metrics used in this work. Section 4 reports and analyzes the experimental results, including comparisons with baseline models and related work. Section 5 discusses the implications of the findings. Finally, Section 6 concludes the article and outlines potential future studies.

## 2. Material and methods

In this section, the materials and methods used in the study are presented. The schematic diagram in Fig. 1 shows the general pipeline of the proposed approach. First, the dataset used for brain tumor classification is introduced, and its scope and imaging details are explained. Next, the use of the EfficientNetV2 architecture as the backbone and the integration of the attention-based MLP-Mixer structure are discussed. Finally, the Grad-CAM-based explainability approach is addressed.

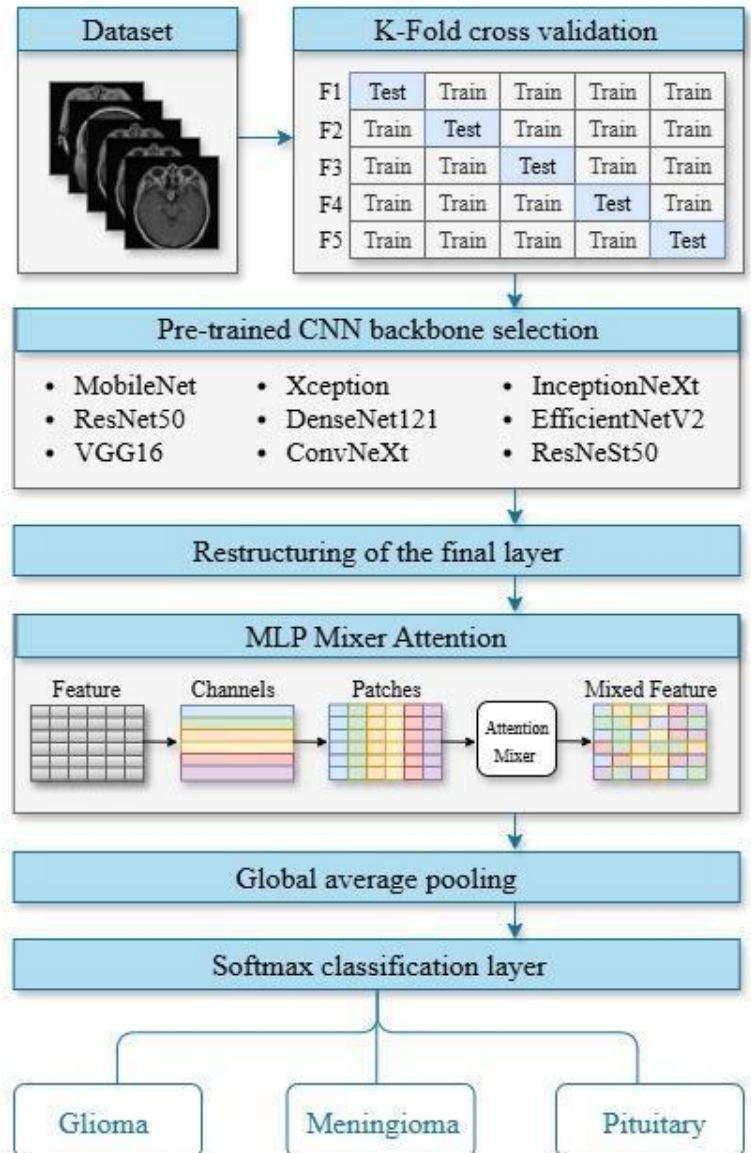

**Figure. 1.** Schematic diagram of proposed study

## 2.1. Dataset

The Figshare dataset[24] used in this study was created from MRI images obtained during brain tumor examinations performed between 2005 and 2010 at Nanfang Hospital and Tianjin Medical University General Hospital in China. It includes 3064 T1-weighted contrast-enhanced brain MRI images from a total of 233 patients. The dataset was first published online in 2015 and was updated with the latest version 8 in 2024, which is the version used in this study. The dataset contains three different tumor types: meningioma (708 images), glioma (1426 images) and pituitary tumor (930 images). Images were acquired in three different planes: sagittal (1025 images), axial (994 images) and coronal (1045 images). Sample images of different tumor types and planes are presented in Fig. 2.

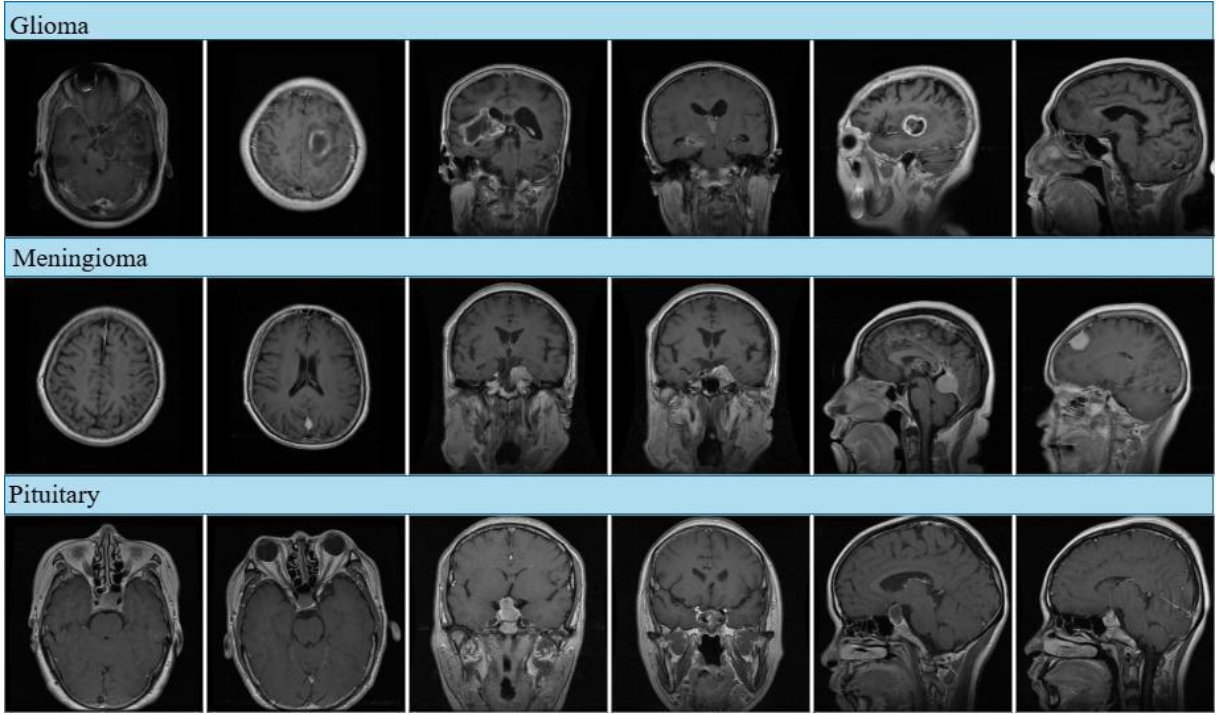

**Figure 2.** MRI slices from the brain tumor dataset for glioma (top), meningioma (middle), and pituitary tumor (bottom). Each row shows axial (first two), coronal (middle two), and sagittal (last two) views.

### 2.2. Linear Attention

Linear attention[25] is a variant of the attention mechanism designed to efficiently capture context dependencies in long sequences with reduced computational cost. The basic principle is to transform the query ($Q$) and key (K) vectors into a positive definite space via a kernel function $\phi(\cdot)$. With that transformation, the attention calculation can be done with linear complexity by rearranging the matrix multiplication order. The general formula is as shown Eq. 1.

$$Attention(Q, K, V) = \frac{\phi(Q)(\phi(K)^T V)}{\phi(Q)(\phi(K)^T 1_n)} \quad (1)$$

Where $V$ is the value matrix and $1_n$ is the unit vector of dimension $n$. The numerator represents the weighted multiplication of queries by values and the denominator represents the normalization of these weights. In Eq. 1, the numerator $\phi(K)^T V$ represents the weighted multiplication of the query vectors with the value matrix, while the denominator $\phi(K)^T 1_n$ corresponds to the normalization of these weights. The kernel $elu(x) + 1$, which produces positive values as a function of $\phi(\cdot)$, was used for this study. This function guarantees a mapping into a positive–definite space, ensuring stable and efficient computation.

### 2.3. MLP Mixer Attention

The MLP Mixer[26] is an architecture that uses a pure multi-layer perceptron (MLP) to perform mixing of information in both the spatial and channel dimensions. In the classic MLP-Mixer, the input data is first passed through Layer Normalization, then interactions between different patches are enabled through the Token-Mixing MLP layer. Subsequently, the feature channels at each location are mixed together using the Channel-Mixing MLP layer. In both stages, residual connections are used to preserve information flow gradient propagation. In this study, a linear attention mechanism has been integrated into the MLP-Mixer structure. In the proposed MLP-Mixer Attention block, an attention module is placed in front of both the Token-Mixing MLP and Channel-Mixing MLP layers. The first attention layer is applied before the Token-Mixing MLP to capture spatial relationships more effectively. The second attention layer is positioned before the Channel-Mixing MLP to strengthen the relationships in the channel dimension. Thus, the pure MLP-based mixing capability of the MLP-Mixer is combined with the long-range relationship capture capacity of the attention mechanism. The proposed MLP-Mixer Attention block is shown in Fig. 3.

The proposed MLP Mixer Attention block can be formulated as follows: in the first stage, the input $X \in R^{nxd}$ is normalized and passed through a spatial attention module $Attn_s$, followed by the Token-Mixing MLP as in Eq. 2.

$$Z' = X + W_2\sigma(W_1 Attention_s(LN(X))) \qquad (2)$$

In the second stage, a channel attention module $Attn_c$ is applied before the Channel-Mixing MLP, as formulated in Eq. 3.

$$Z'' = Z' + V_2\sigma(V_1 Attention_c(LN(Z'))) \qquad (3)$$

In Eq. 2 and 3, $X \in R^{nxd}$ denotes the input matrix with $LN(\cdot)$ represents Layer Normalization, while $Attention_s$ and $Attention_c$ denote the token and channel attention modules, respectively. $W_1$ and $W_2$ correspond to the learnable parameters of the Token-Mixing MLP, $V_1$ and $V_2$ are the parameters of the Channel–Mixing MLP. $\sigma(\cdot)$ is a non–linear activation function, and the residual connections preserve the information flow.

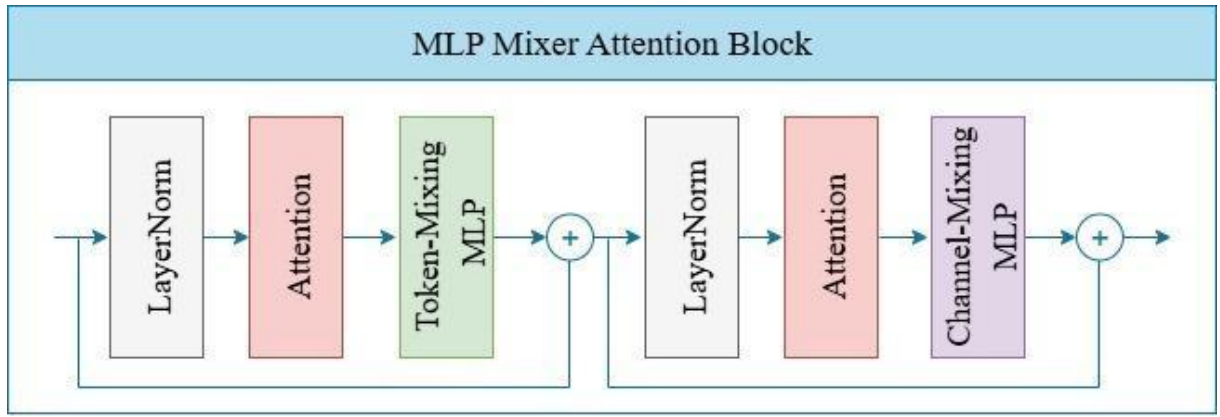

**Figure 3.** Schematic representation of the proposed MLP-Mixer Attention block

## 2.4. Proposed methodology

EfficientNetV2[27] was selected as the backbone of the model because it provides an optimal balance between high accuracy and computational efficiency. EfficientNetV2 has different scales (depth, width, and resolution). A small scale was selected for this study. Fig. 4 shows an overview of the proposed EfficientNetV2–MLP Mixer Attention model. Input images are first passed through a standard *3 × 3* convolution layer to extract low-level edge, texture, and simple shape information. In the next step, these extracted features are processed sequentially through fused MBConv and MBConv blocks. Fused

MBConv blocks consist of a *3 × 3* convolution, a squeeze and excitation (SE) module, and a *1 × 1* convolution with residual connections. Fused MBConv provides stronger feature extraction while keeping computational costs low, especially in low-resolution first layers. In contrast, classic MBConv blocks contain a *3 × 3* convolution in the depth direction between two *1 × 1* convolutions and an SE module. These blocks significantly reduce the number of parameters while enabling the capture of more detailed spatial relationships. Features extracted from the backbone are first downsampled using a *1 × 1* convolution, Then, this tensor of dimensions $H \times W \times d$ is reshaped to the form $R^{n \times d}$, where $n = H \times W$, and each spatial location is represented as a patch. The resulting feature vector is then fed into the proposed MLP-Mixer Attention block. In the final stage, global average pooling (GAP) is applied, and the final classification results are obtained using a fully connected layer with softmax activation.

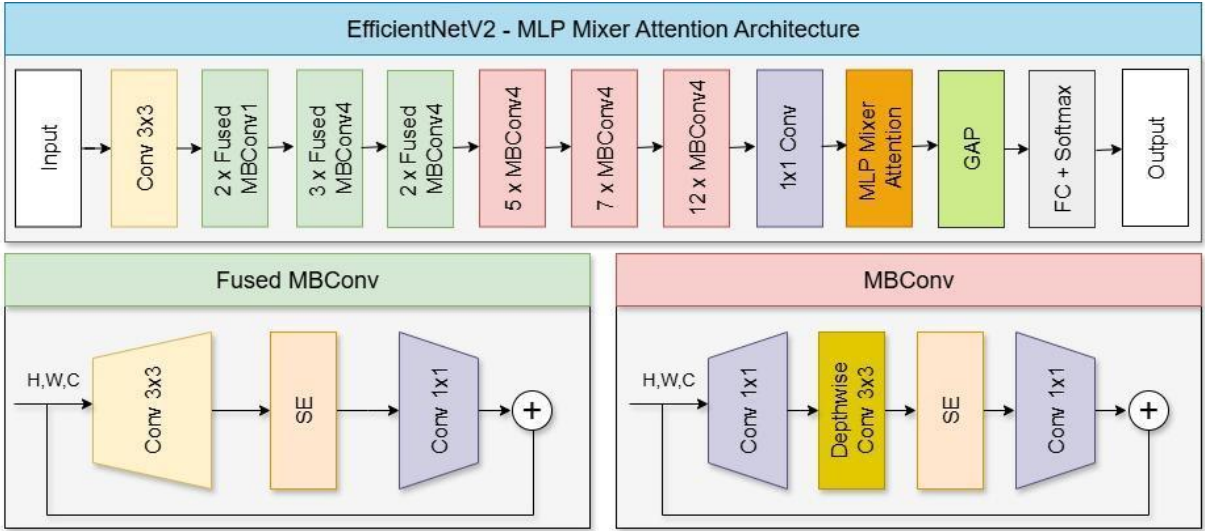

**Figure. 4.** Schematic representation of the proposed EfficientNetV2–MLP Mixer Attention model

## 2. 5. Explainability with Grad-CAM

Grad-CAM[28] is a method used to visualize how DL models make predictions. It aims to find out which areas of the image the model pays more attention to when making decisions. First, the score function $y^c$ of the target class is calculated, and the derivatives of this score on each feature map in the final convolution layer is calculated on $A^k$ . The average of the derivatives is then used to find the weight ($a_k^c$) that shows how important each feature map is for the class.

$$a_k^c = \frac{1}{Z} \sum_i \sum_j \frac{\partial y^c}{\partial A_{ij}^k} \tag{4}$$

As shown in Eq. 4, the $a_k^c$ coefficient indicates how much importance the model places on the $k^{th}$ feature map when making class predictions. In the final stage, a heat map is created using the obtained weights.

$$L_{Grad-CAM}^c = ReLU(\sum_k a_k^c A^k) \tag{5}$$

Eq. 2 produces the Grad-CAM map. The reason we used ReLU here is that we only want to show the regions that contribute positively to the class prediction. As a result, the resulting heat map reveals which regions of the image are influential in the model's decision.

## 3. Experimental Setup

The experiments in this study were performed using a workstation with the hardware and software specifications listed in Table 1. In all experiments, the data set was evaluated using a 5-fold cross-validation method. The reason for choosing the number of folds as 5 during cross-validation is that the data set used is generally preferred as 5-fold in the literature, as well as to allow a fair comparison of the methods.

**Table 2.** Hardware and software specifications of the workstation used in the experiments.

| System Component | Specification |
| --- | --- |
| CPU | Intel® Core™ i9-10920X CPU @ 3.50GHz |
| RAM | 128 GB |
| GPU(s) | 2 × NVIDIA RTX A5500 (24 GB each) |
| Operating System | Windows 10 64 bit |
| Python Version | 3.9.23 |
| Keras Version | 2.10 |

### 3.1. Performance Evaluation Metrics

The objective evaluation of deep learning model performance is crucial. In this study, accuracy, precision, recall and F1-score commonly used in the literature are utilized. These metrics are calculated based on the complexity matrix created by comparing the model's predictions with the actual labels. Accuracy is the overall correctness, precision is the accuracy of the model's positive predictions, recall is how many true positives are correctly predicted, and F1-score is a balanced combination of precision and recall. The formulas of the performance evaluation metrics are presented in Eq. 6 to 9.

$$Accuracy = (TP + TN)/(TP + TN + FP + FN) \tag{6}$$

$$Precision = TP/(TP + FP) \tag{7}$$

$$Recall = TP/(TP + FN) \tag{8}$$

$$F1 - Score = 2(Precision \; x \; Recall)/(Precision + Recall) \tag{9}$$

## 4. Results

In this section, the results of all experiments conducted within the scope of the study are analyzed. The five-fold cross-validation method was used in all experiments. In order to objectively evaluate the model performance, the mean values of the cross-validation results were calculated.

### 4.1. Baseline model results

Within the scope of the study, nine different CNN architectures commonly used in the literature were tested to determine the backbone model. Table 3 provides the performance of these models in brain tumor classification in detail. When the table is examined, it is seen that the models achieved accuracy rates in the range of 93–97%. Furthermore, the fact that the precision, recall, and F1-score values are consistent with and close to the accuracy value indicates that the models demonstrate a balanced performance not only in terms of overall accuracy but also in terms of reducing false positives and capturing true positives.

As a result of the experimental study, the most successful model was EfficientNetV2, with 97.54% accuracy, 97.93% precision, 96.49% recall, and 97.11% F1-score values. EffectiveNetV2's success can be attributed to the advantages provided by its scaling approach and the more efficient feature extraction enabled by its fused MBConv blocks. It enabled more effective learning of both low-level and high-level spatial relationships, thereby improving classification performance. Meanwhile, models such as VGG16 (96.88% accuracy, 96.40% precision, 96.38% recall, 96.36% F1) and DenseNet121 (96.21% accuracy, 96.03% precision, 95.65% recall, 95.72% F1) also showed high and balanced performance. It is particularly noteworthy that VGG16 still has strong representational capability despite its simple layer structure. MobileNet, which has a lighter architecture, combines computational efficiency with high accuracy, achieving 95.98% accuracy, 95.86% precision, 95.08% recall, and 95.29% F1-score.

The InceptionNeXt model achieved balanced but medium-level performance with 94.77% accuracy, 93.91% precision, 94.64% recall, and 94.17% F1-score values. ResNeSt50 achieved 94.12% accuracy, 94.01% precision, 92.37% recall, and 92.84% F1-score, but due to the low recall value, it missed some of the positive examples. Xception, on the other hand, demonstrated a stable and balanced performance with 95.82% accuracy, 95.34% precision, 95.03% recall, and 95.12% F1-score.

In contrast, the ConvNeXt model showed limited success in detecting tumor cases, with accuracy, precision, recall, and F1-score values of 93.95%, 94.25%, 91.90%, and 92.72%, respectively, due to its relatively low recall rate. ResNet50, on the other hand, demonstrated the lowest performance with 93.43% accuracy, 93.34% precision, 91.70% recall, and 92.15% F1-score. These results reveal that, despite acceptable precision values in both models, a portion of positive classes were missed due to low recall. Therefore, classic ResNet architectures and structures such as ConvNeXt are not as successful as modern and optimized architectures in distinguishing complex tumor patterns.

**Table 3.** Brain tumor classification performances of baseline CNN models

| Model | Performance measurement metrics(%) | | | |
|---|---|---|---|---|
| | Accuracy | Precision | Recall | F1 Score |
| ConvNeXt[29] | 93.95 | 94.25 | 91.90 | 92.72 |
| DenseNet121[30] | 96.21 | 96.03 | 95.65 | 95.72 |
| **EfficientNetV2[27]** | **97.54** | **97.93** | **96.49** | **97.11** |
| InceptionNeXt[31] | 94.77 | 93.91 | 94.64 | 94.17 |
| MobileNet[32] | 95.98 | 95.86 | 95.08 | 95.29 |
| ResNeSt50[33] | 94.12 | 94.01 | 92.37 | 92.84 |
| ResNet50[34] | 93.43 | 93.34 | 91.70 | 92.15 |
| VGG16[35] | 96.88 | 96.40 | 96.38 | 96.36 |
| Xception[36] | 95.82 | 95.34 | 95.03 | 95.12 |

The numerical results presented in Table 3 are also compared visually in Fig. 5. The figure presents the differences between the models' accuracy, precision, recall, and F1-score values in a more comprehensive manner, clearly reflecting the impact of the decline in recall values and the emergence of balanced architectures.

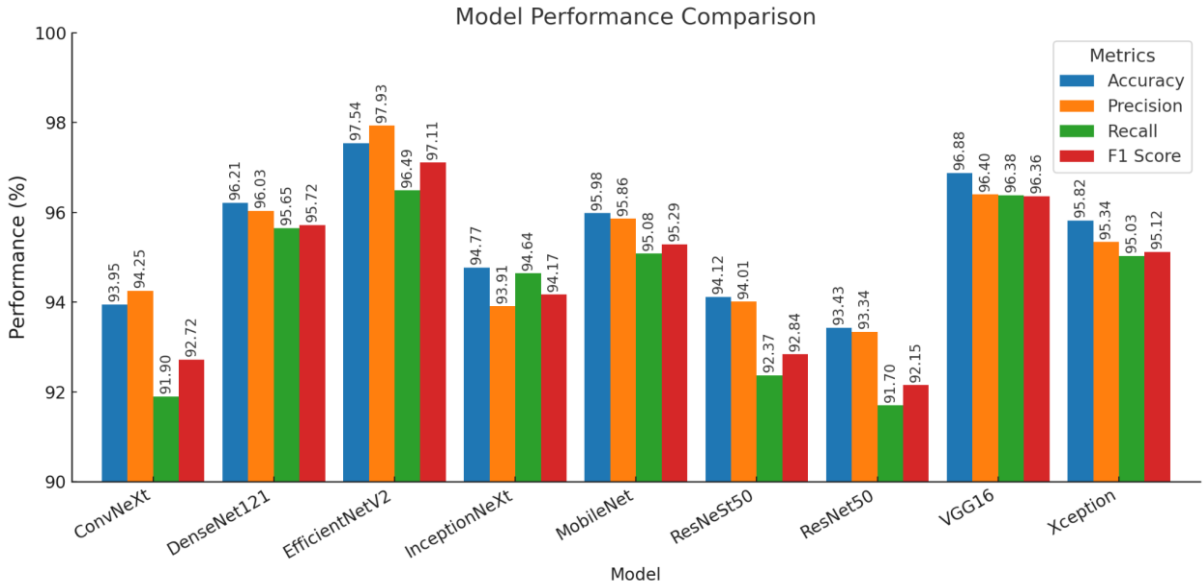

**Figure 5.** Comparison of DL models across performance metrics

In conclusion, the findings reveal that CNN-based architectures provide a robust framework for brain tumor classification. The comparison between the models shows that EfficientNetV2 stands out in particular for achieving balanced results in terms of both high accuracy and other metrics such as precision, recall, and F1-score. Therefore, the preference for EfficientNetV2 as the backbone architecture is one of the most important outcomes of this experimental study.

### 4.2. Proposed model results

This section presents detailed experimental results obtained by the hybrid model based on EfficientNetV2 and enhanced with MLP-Mixer Attention blocks. Table 4 shows the performance metrics of experiments conducted with different token and channel sizes.

**Table 4.** Performance comparison of the proposed EfficientNetV2–MLP Mixer Attention model with different token and channel dimensions.

| EfficientNetV2-MLP Mixer Attention Dimension | | Performance measurement metrics(%) | | | |
|---|---|---|---|---|---|
| **Token** | **Channel** | **Accuracy** | **Precision** | **Recall** | **F1 Score** |
| 64 | 256 | 98.53 | 98.66 | 97.88 | 98.23 |
| **128** | **512** | **99.50** | **99.47** | **99.52** | **99.49** |
| 256 | 1024 | 98.85 | 98.65 | 98.76 | 98.71 |

When the results in the table are examined in detail, it can be seen that the model achieves a very high success rate of over 98% in all configurations. In particular, the model with Token = 128 and Channel = 512 dimensions achieved the best performance, reaching 99.50% accuracy, 99.47% precision, 99.52% sensitivity, and 99.49% F1 score. These values demonstrate that the model not only maximizes the correct classification rate but also achieves a balanced success by minimizing false positive and false negative rates. When the performance of the optimal configuration (Token = 128, Channel = 512) is compared with the baseline EfficientNetV2 model, the proposed approach demonstrates a relative improvement of 1.96% in accuracy (from 97.54% to 99.50%), 1.54% in precision (from 97.93% to 99.47%), 3.03% in recall (from 96.49% to 99.52%), and 2.45% in F1-score (from 97.11% to 99.49%). Although these improvements may appear numerically small, in the context of healthcare applications, even a marginal increase of less than 0.1% can be clinically significant. Therefore, the performance gains obtained in this study are of considerable value for reliable and effective AI-based decision support systems in medical imaging.

Although high success rates were observed in the other two configurations, it was noted that the sensitivity rate (97.88%) remained relatively low in the smaller 64–256 dimension, resulting in a small portion of tumor cases being missed. In contrast, in the 256–1024 dimension, despite the additional computational cost associated with the larger dimension, it is noteworthy that the performance remains at a lower level compared to the medium configuration. This situation shows that as the dimension increases in the model, overfitting or unnecessary parameter redundancy may negatively affect performance. Therefore, the 128–512 configuration, which provides the best balance, can be considered the optimal setting for the model.

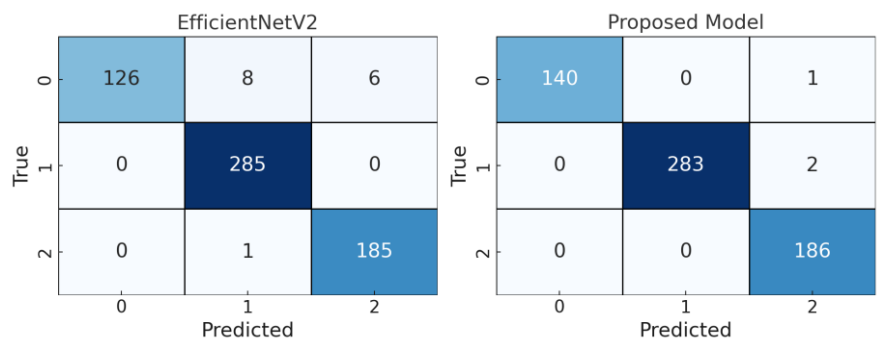

**Figure 7.** Confusion matrices of the EfficientNetV2 and proposed model(0: meningioma, 1: glioma, 2: pituitary tumor)

The confusion matrices presented in Fig. 7 show the classification performance of the basic EfficientNetV2 model and the proposed model. In the EfficientNetV2 model, misclassifications are particularly observed between the meningioma and glioma classes (8 glioma cases were classified as meningioma, and 6 cases were classified as pituitary tumors). In contrast, in the proposed model, only three misclassifications occurred across tumor types, and errors were significantly reduced. Thus, the integration of MLP-Mixer Attention has increased the model's discriminatory capability, improving overall accuracy and largely eliminating uncertainty between classes.

### 4.3. Grad-CAM Results

The Figshare dataset[24] contains the tumor masks for each MR slice, and these masks clearly show the region where the tumor is located. Therefore, it is possible to analyze the regions that the proposed model focuses on and evaluate the clinical accuracy of the model's prediction processes. Fig. 8 shows example images from different planes (axial, coronal, and sagittal) and different tumor types (glioma, meningioma, and pituitary tumor). The first column shows the original T1-weighted contrast-enhanced MRI slices, the second column shows the tumor regions marked by experts (ground truth), and the third column shows the focus regions highlighted by the proposed model using the Grad-CAM method.

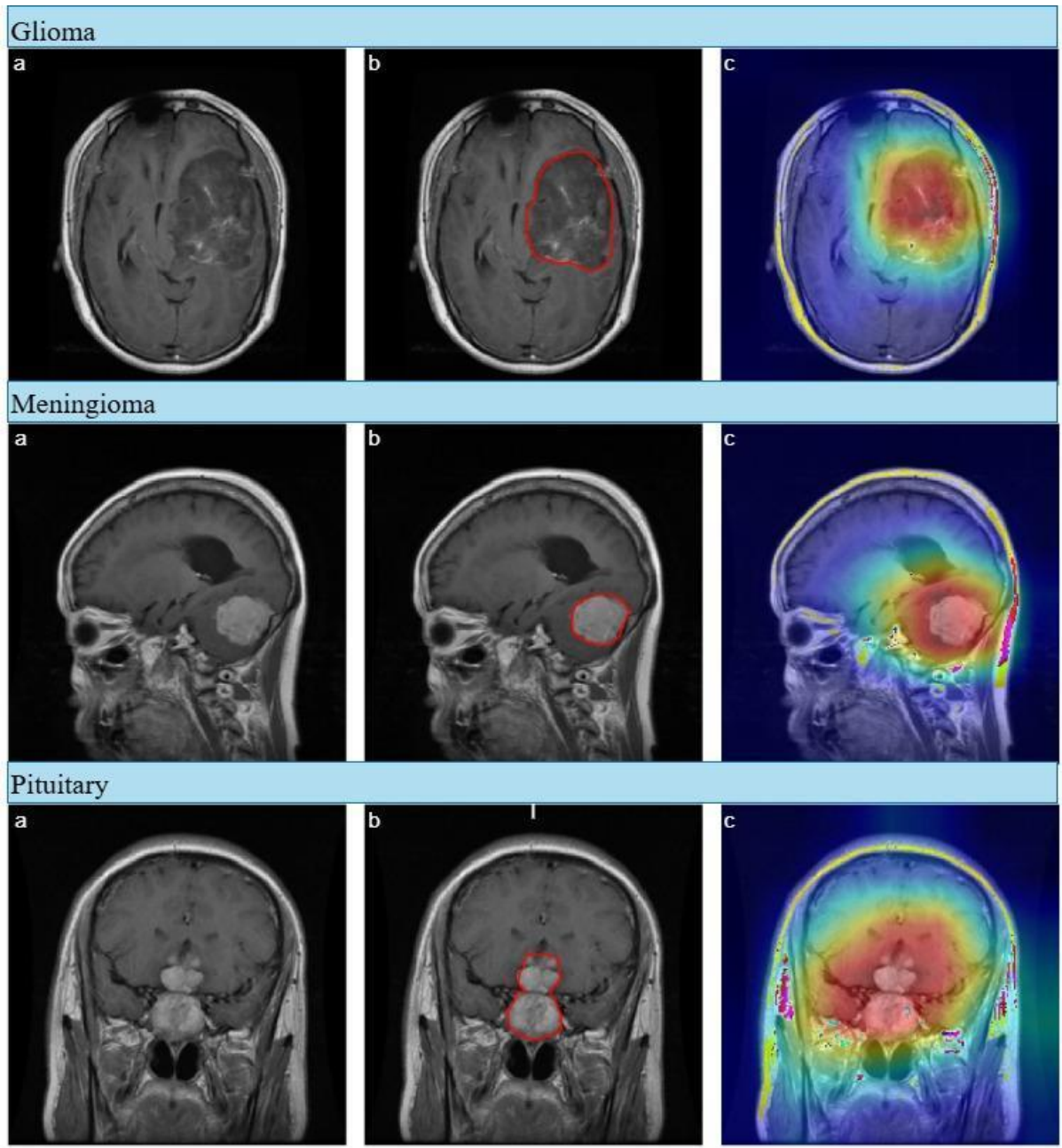

**Figure 8.** Representative MRI slices of brain tumors: (a) original T1-weighted contrast-enhanced images, (b) tumor regions annotated with ground truth, and (c) Grad-CAM heatmaps highlighting the model's focus areas for classification in glioma, meningioma, and pituitary tumors.

When Fig. 8 is examined, it can be seen that the areas on which the model focuses intensively largely match with real tumor regions. In glioma samples, the model focuses on covering the entire extensive lesions that show heterogeneous structures. In meningioma cases, it is seen that the model even shows sensitivity to the tissue boundaries surrounding the tumor. In pituitary tumors, even small and difficult-to-distinguish lesion areas are correctly identified and targeted.

As a result, Grad-CAM visualizations enhance the explainability of the proposed hybrid model and enable clinicians to use AI-based systems more reliably. It is particularly important to be able to show which areas the model focuses on in order to counter "black box" criticism, especially in the field of healthcare. These findings demonstrate that the model not only achieves high classification accuracy but also makes decisions by considering clinically meaningful regions.

### 4.4. Comparison with literature studies

The performance of the proposed model in Table 5 has been compared with recent studies in the literature. As can be seen, the study by Hashemzehi et al. (2020) achieved limited success with an accuracy and F1 score of around 94%. Although the accuracy rates reported in the studies by Swati et al. (2019) and Ismael et al. (2020) were 96.49% and 97%, respectively, the precision and sensitivity values were lower than the proposed model. Aamir et al. (2022) achieved high success with 98.95% accuracy; however, other performance metrics were not reported, and the computational cost is high due to the multi-stage structure of the model. In Davar et al. (2025), sensitivity remained at 89.74%, resulting in relatively low performance.

**Table 5.** Comparison of the performance result of the proposed model with literature studies

| Study | Performance measurement metrics(%) | | | |
|---|---|---|---|---|
| | **Accuracy** | **Precision** | **Recall** | **F1- Score** |
| Hashemzehi et al. (2020) | - | 94.49% | 94.64% | 94.56% |
| Swati et al. | 96.49 | 92.79 | 96.33 | 94.47 |
| Aamir et al. | 98.95 | - | - | - |
| Ismael et al. | 97 | 98 | 97 | 97 |
| Davar et al. | - | 91.27% | 89.74% | - |
| **Proposed model** | **99.50** | **99.47** | **99.52** | **99.49** |

The proposed hybrid EfficientNetV2–MLP Mixer Attention model demonstrates higher and more balanced results than all other studies, with 99.50% accuracy, 99.47% precision, 99.52% sensitivity, and 99.49% F1 score. The superiority achieved in both sensitivity and specificity metrics demonstrates that the model minimizes both false positive and false negative classifications. Considering that even one-tenth of a percent improvements are clinically critical in the field of healthcare, these improvements prove that the proposed method makes a significant contribution to the literature.

### 5. Discussion

Brain tumors are serious health problems caused by the uncontrolled and abnormal proliferation of cells in the brain and have high mortality rates. Therefore, early diagnosis and accurate classification play a critical role in patient treatment. The EfficientNetV2–MLP Mixer Attention-based method proposed in this study offers a powerful artificial intelligence approach to the classification of brain tumors. The study was conducted in two stages. In the first stage, nine different CNN architectures commonly used in the literature were tested to determine the most suitable backbone. As shown in Table 3, EfficientNetV2 was the most successful model with an accuracy rate of 97.54%, while ResNet50 showed the lowest performance with an accuracy of 93.43%. As a result of this evaluation, EfficientNetV2 was selected as a powerful feature extractor.

In the second stage, EfficientNetV2 was combined with the proposed attention-based MLP-Mixer structure. The results obtained are presented in Table 4. In particular, with the Token = 128 and Channel = 512 configuration, the model achieved the highest performance with 99.50% accuracy, 99.47% precision, 99.52% sensitivity, and 99.49% F1 score. Compared to the basic EfficientNetV2, these values represent an increase of 1.96% in accuracy, 1.54% in precision, 3.03% in sensitivity, and 2.45% in F1 score. Considering that even an improvement of one thousandth in error rates is of vital importance in clinical applications, these improvements are extremely valuable. The model's complexity matrix in Fig. 7 was used to evaluate the error distribution. While EfficientNetV2 showed notable misclassifications between meningioma and glioma classes, these errors were significantly reduced in the hybrid model. This finding indicates that the proposed approach achieves clearer inter-class differentiation. The model's explainability was examined using the Grad-CAM visualizations provided in Fig. 8. In examples of glioma, meningioma, and pituitary tumors, the regions of interest identified by the model largely overlap with the actual tumor masks. This result demonstrates that the model makes its predictions based on clinically meaningful regions.

Therefore, the proposed model provides a strong contribution in terms of explainability as well as high performance. The findings obtained also show superiority when compared with current studies in the literature. As summarized in Table 5, Swati et al. (2019) reported 96.49% accuracy, Ismael et al. (2020) reported 97% accuracy, and Aamir et al. (2022) reported 98.95% accuracy. However, in some of these studies, explainability was limited, and in others, the computational cost was high. The proposed model achieved the highest success reported in the literature with 99.50% accuracy and provided balanced superiority in all metrics. However, the study also has some limitations. The Figshare dataset used only includes three tumor types and does not cover the different tumor types that may be encountered in real clinical conditions. Additionally, the fact that it has not been tested in different imaging modalities limits the model's generalization ability. Future studies plan to overcome these limitations by collaborating with healthcare institutions to collect more comprehensive and diverse datasets, reevaluating the model with different imaging techniques, and therefore addressing the current limitations.

## 6. Conclusion
This paper proposes an EfficientNetV2-based and Attention-supported MLP-Mixer architecture for the classification of brain tumors. The model has surpassed existing methods by achieving high accuracy, precision, sensitivity, and F1 scores in comprehensive experiments conducted on the Figshare dataset. In particular, the results achieved with the optimal configuration demonstrate that the developed approach has strong feature extraction and classification capabilities. Grad-CAM visualizations revealed that the regions of interest identified by the model corresponded to actual tumor areas, therefore providing a significant advantage not only in terms of performance but also in terms of clinical reliability. This increases the method's usability as a decision support tool for radiologists. The results once again demonstrate that artificial intelligence-based systems can provide reliable solutions in the healthcare field. Future studies using larger, multi-center, and multimodal datasets will increase the method's generalization capacity and enable its more widespread use in clinical applications.

**CRediT authorship contribution statement**
Mustafa Yurdakul and Şakir Taşdemir conducted the entire conceptualization, experimentation, analysis, interpretation, writing, and revision of the study independently.

**Declaration of competing interest**

The author declares no competing interests.

**Data availability**

Figshare brain tumor dataset is publicly available at the following link.
https://figshare.com/articles/dataset/brain_tumor_dataset/1512427